\documentclass[prx,twocolumn,superscriptaddress,aps,longbibliography,10pt]{revtex4-2}

\usepackage{mathtools}  

\usepackage{amssymb}

\usepackage{natbib}

\usepackage{graphicx}

\usepackage{float}

\usepackage{enumitem}

\usepackage[dvipsnames]{xcolor}

\usepackage[pdftex,colorlinks=true,urlcolor=blue,linkcolor=blue,citecolor=orange,breaklinks=true,bookmarks=false]{hyperref}

\usepackage{tikz}

\usepackage{url}            

\usepackage{booktabs}       

\usepackage{nicefrac}       

\usepackage{microtype}      

\usepackage{lipsum}

\graphicspath{{figs/}{.}}     

\definecolor{violet}{rgb}{0.56, 0, 1}

\begin{document}

\title{Tabular foundation models for non-tabular tasks}

\author{Goran Nakerst}
\affiliation{Institut f\"ur Theoretische Physik, Technische Universit\"at Dresden, 01062 Dresden, Germany}

\author{John Brennan}
\affiliation{Department of Physics, Maynooth University, Ireland}

\author{Wouter Beugeling}
\affiliation{Physikalisches Institut (EP3) and Institute for Topological Insulators, Universit\"at W\"urzburg,
Am Hubland, 97074 W\"urzburg, Germany}

\author{Masudul Haque}
\affiliation{Institut f\"ur Theoretische Physik, Technische Universit\"at Dresden, 01062 Dresden, Germany}


\begin{abstract}
Tabular foundation models (TFMs) have recently emerged as a promising paradigm for machine learning on tabular data, offering the ability to generalize across datasets without task-specific training. Since many machine learning datasets can be represented as tables, this raises the question: does TFM capability extend beyond tasks traditionally regarded as tabular?  We address this question by using TabPFN v3 on three non-tabular classification problems: handwritten digit recognition on MNIST, language identification of French and German words, and image classification on Tiny ImageNet. In each case, the original data are represented as rows of a table and classification is formulated as prediction of a missing label.  We evaluate performance as a function of the number of context samples provided to the pretrained model, with no additional training or fine-tuning. Despite having no explicit access to the spatial or sequential structure characterizing the data, TabPFN v3 in some cases achieves accuracies comparable with that of models or methods geared specifically toward the corresponding tasks.

\end{abstract}


\maketitle

\section{Introduction}

Tabular data is arguably the dominant format in applied machine learning.  For over a decade, gradient-boosted decision trees, exemplified by XGBoost \cite{chen2016xgboost} and LightGBM \cite{ke2017lightgbm}, have been the dominant tools for such data. 
These methods typically involve training each model from scratch on a single type of dataset, thus their capacity to transfer knowledge across tasks or datasets is limited.  In contrast, in other machine learning fields such as language and vision, the idea of `foundation' models has become central \cite{Devlin_Toutanova_2018_BERT, Brown_EtAl_2020:languagemodelsfewshotlearners, Dosovitskiy_EtAl_2020_ImageWorth16x16Words, RadfordEtAl_Sutskever_2021_TransferableVisualModels, Bommasani_EtAl_2022_opportunitiesrisksfoundationmodels} --- pre-training models on a large corpus  to achieve impressive generalization capabilities across domains or tasks.  Thus, perhaps unsurprisingly, the idea of importing this paradigm of foundation models to tabular data has gained appeal, and  tabular foundation models (TFMs) have experienced an upsurge of interest over the last few years \cite{VanBreugel_VanDerSchaar_2024_WhyTFMPriority, hollmann2025tabpfn, grinsztajn2025tabpfn_25, grinsztajn2026tabpfn_3, qu2025tabicl, Sergazinov_Yin_2025_ChunkedTabPFN, qu2026tabiclv2, jiang2026tfmsurvey, LiuEtAl_2026_GraphAnomalyDetectors, Luo_Leontjeva_Lucey_2026_MemoryEfficientTFMs, Kloetergens_Hanika_2026_TFMsAgree}.  In particular, a few such models have now become available for public use, most prominently TabPFN \cite{hollmann2025tabpfn, grinsztajn2025tabpfn_25, grinsztajn2026tabpfn_3}, TabICL \cite{qu2025tabicl, qu2026tabiclv2} and recently TabFM \cite{google2026tabfm}. 

With the increasing prominence of tabular foundation models, a  very natural question arises: Since most datasets can be represented in tabular form, perhaps TFMs are applicable to many machine learning tasks, beyond those conventionally regarded as tabular?   As an example, we show schematically in Figure \ref{tab:mnist_as_table} how the MNIST dataset could be thought of as a tabular dataset.  Each complete row of the table represents one sample, with $28\times28=784$ entries containing the pixel values and the last entry containing the label, a single-digit integer.  The schematic shows the case with two learning samples per digit.  These twenty rows would serve the training set for a machine learning algorithm, or the context set  (or support set) for a pre-trained foundation model.  Test samples correspond to rows in which the label cell is not filled in; inference involves predicting the content of this cell.  Learning image classification is thus recast into the task of learning to fill in missing values in a table, i.e., the task that tabular foundation models are built for.

\begin{figure}[tbp]
\centering
\begin{tabular}{|c|c|c|c|c|c|}
\hline
Image sample & \multicolumn{4}{|c|}{Pixel values} & Label \\
\hline \hline
1 & $p_1$(1) & $p_1$(2) & \phantom{xxx}\ldots\phantom{xxx}\ldots\phantom{xxx} & $p_1$(784) & 0 \\
\hline
2 & $p_2$(1) & $p_2$(2) & \phantom{xxx}\ldots\phantom{xxx}\ldots\phantom{xxx} & $p_2$(784) & 0 \\
\hline
3 & $p_3$(1) & $p_3$(2) & \phantom{xxx}\ldots\phantom{xxx}\ldots\phantom{xxx} & $p_3$(784) & 1 \\
\hline
4 & $p_4$(1) & $p_4$(2) & \phantom{xxx}\ldots\phantom{xxx}\ldots\phantom{xxx} & $p_4$(784) & 1 \\
\hline
\vdots & \vdots & \vdots & \vdots & \vdots & \vdots \\
\hline
20 & $p_{20}$(1) & $p_{20}$(2) & \phantom{xxx}\ldots\phantom{xxx}\ldots\phantom{xxx} & $p_{20}$(784) & 9 \\
\hline \hline
q1 & $p_\text{q1}^{}$(1) & $p_\text{q1}^{}$(2) & \phantom{xxx}\ldots\phantom{xxx}\ldots\phantom{xxx} & $p_\text{q1}$(784) & ?? \\
\hline
q2 & $p_\text{q2}^{}$(1) & $p_\text{q2}^{}$(2) & \phantom{xxx}\ldots\phantom{xxx}\ldots\phantom{xxx} & $p_\text{q2}$(784) & ?? \\
\hline
\end{tabular}
\caption{Framing MNIST classification as a tabular task.  In this schematic, two examples per MNIST digit, thus a total of 20 images, are used as the context (the support set) from which the TFM learns.  The last two rows, labeled q1 and q2 (q for 'query'), are images to be used for inference/prediction; hence the Label field is empty.  The collection of unlabeled samples, the `test' data set, is sometimes called  the `query set' or 'target set' when the learning is in-context.}
\label{tab:mnist_as_table}
\end{figure}

In this work, we ask to what extent currently available TFMs are suitable for tasks which are not usually regarded as tabular.  We present empirical results for three tasks:  handwritten digit recognition using MNIST, differentiating between french and german words, and a 100-class Tiny ImageNet dataset.  We use TabPFN v3 \cite{hollmann2025tabpfn, grinsztajn2025tabpfn_25, grinsztajn2026tabpfn_3} and test the performance with various numbers of  samples used as the `context'.  Since the model is pre-trained, there is no training involved in the sense of updating model weights; instead, learning is entirely \emph{in-context}.    We compare classification accuracies with those  obtained by \emph{training} more specialized models, such as convolutional neural networks, on the same number of samples.  In some cases, the performance is of similar order, despite the fact that the TFM undergoes no task-specific training.

These results indicate that the capabilities of current TFM are not necessarily confined to conventionally tabular data, for which they were originally designed.  We thus expect that it might be beneficial to reconsider the distinction between tabular and non-tabular learning.  Since TabPFN is trained on causal graphs, this work also suggests that (pre-)training for understanding causality may be an important key to broad generalization across task types. 

We present performance comparisons between foundation and non-foundation models.  The comparisons are done such that the size of the support set for in-context learning of foundation models is the same as the size of the training set for training of non-foundation models.  This makes the terminology potentially confusing, as the terms `training' and `test' are commonly used for foundation models as well.  We hope that the distinction between in-context learning on the one hand (for TFMs), and training with model weight updating on the other hand (for other techniques), is clear in every case that we discuss in this article, even if the same word (`training') is sometimes used for both types of learning.

In Section \ref{sec:TFMreview} we provide a brief reminder of the basics of tabular foundation models; readers familiar with TFMs and tabular data may safely skip this section.  The next three sections present the results on MNIST (Section \ref{sec:mnist}), on the language recognition task  (Section \ref{sec:nlp}) and on the Tiny ImageNet dataset (Section \ref{sec:tiny_imagenet}).  Section \ref{sec:conclusion}  provides a summary and some discussion, and the appendices provide more technical information regarding the numerical experiments.

\section{Basics of Tabular Foundation Models}

\label{sec:TFMreview}


Tabular machine learning is commonly used for the prediction of target columns: given a table where values for a subset of feature columns are known, the objective is to predict missing values in a target column. Given that machine learning methods in other domains, like image, audio and language processing, have been dominated by neural network models in recent decades, one might expect deep learning to have had a similar impact in tabular data tasks. However, until recently, breakthrough deep learning models have remained elusive in this domain, despite some efforts \cite{BorisovEtAl_2024_DNNsTabularDataSurvey, Gorishniy2021, Arik2021, Popov2019, Huang2020, Somepalli2021, Song2018}.
This has been attributed to several reasons, e.g., (1) Tabular data typically lacks the spatial or sequential structures that deep learning architectures like CNNs natively exploit.  (2) Tabular data is typically heterogeneous in terms of type, such as numerical, categorical, and text data, and may contain missing values. While classical ensemble models can handle missing data and discrete categories naturally, neural networks require careful encoding strategies to map these features into a vector space without introducing bias.  (3) Tabular data possess a high degree of symmetry of a type that is not natively exploited by many deep learning models, namely, tables are usually invariant under permutation of either rows or columns. This is a consequence of columns/rows only carrying semantic meaning based on their identity rather than their position in the table, so that swapping columns/rows does not alter the underlying data representation. 





Possibly due to these reasons, traditional machine learning methods like gradient-boosted decision trees (e.g., XGBoost, LightGBM, CatBoost and random forests \cite{chen2016xgboost, ke2017lightgbm, prokhorenkova2018catboost, breiman2001random}) have remained the standard approach for tabular tasks in industry \cite{grinsztajn2022why}.
However, recent advances in TFMs have challenged this paradigm \cite{hollmann2025tabpfn, grinsztajn2025tabpfn_25, grinsztajn2026tabpfn_3, qu2025tabicl, qu2026tabiclv2, jiang2026tfmsurvey}. 
A TFM is a single, pre-trained neural network for performing predictive tasks on tabular datasets it did not see during training. By using a transformer-based architecture, trained on a large collection of synthetic datasets generated via causal graphs, TFMs learn abstract patterns of causal relationships between features that generalize across tables. This enables them to match or exceed state-of-the-art classical methods through in-context learning: loading both a set of context/support rows (with known target labels) and query rows (where the target label is missing) directly into the transformer's context window to produce predictions in a single forward pass. Consequently, TFMs can be applied to unseen tasks without requiring possibly costly retraining or hyperparameter tuning, unlike traditional methods. Currently, models like TabPFN \cite{hollmann2025tabpfn, grinsztajn2025tabpfn_25, grinsztajn2026tabpfn_3} and TabICL \cite{qu2025tabicl, qu2026tabiclv2} make TFMs a new default approach for small- to medium-sized tabular prediction tasks.

To process tabular data without flattening it into a single sequence, TFMs use a transformer architecture designed to understand two-dimensional grid structures. Early versions of TabPFN achieve this by introducing an architecture that uses a stack of modules consisting of two attention mechanisms. The first applies an attention layer across columns to capture dependencies between features. The second applies an attention layer across rows of the table to capture relationships between different samples. Alternating row and column attention layers come with two key advantages over using a standard attention mechanism on a flattened table. Firstly, it allows the model to scale efficiently with the table size: for a table with $N$ rows and $M$ columns, a standard transformer scales as $(NM)^2$, while alternating row and column attention scales as $NM^2 + N^2M$. Secondly, alternating row and column attention explicitly preserves invariance under both row and column permutation.

Despite these architectural advantages, processing tabular data via in-context learning introduces some constraints regarding dataset scalability. As the entire support dataset must fit into the transformer's context window alongside query points, the maximum dataset size is strictly bounded by the size of the context window. Consequently, earlier TFM implementations are restricted to relatively small tables (e.g., up to 10,000 rows and 500 features). While recent iterations like TabPFN v3 significantly expand these bounds through feature subsampling and ensembling, a trade-off remains between feature capacity and row capacity. These context limits mean that for large-scale enterprise datasets with millions of rows or tens of thousands of features, classical methods like gradient-boosted decision trees remain indispensable.

In addition to pre-training on tabular data, TFMs are also pre-trained on a large collection of synthetic datasets using structural causal models, which represent causal relationships in the data via a directed acyclic graph.  (Versions of TabPFN and TabICL differ in details of how this idea is implemented, extended or modified in various ways.)  This endows the TFMs with an abstract understanding of the concept of causal relationships, which suggests the potential for generalization beyond their original intended domain.

\section{MNIST}
\label{sec:mnist}

\begin{figure}[tbp]
\includegraphics[width=\columnwidth]{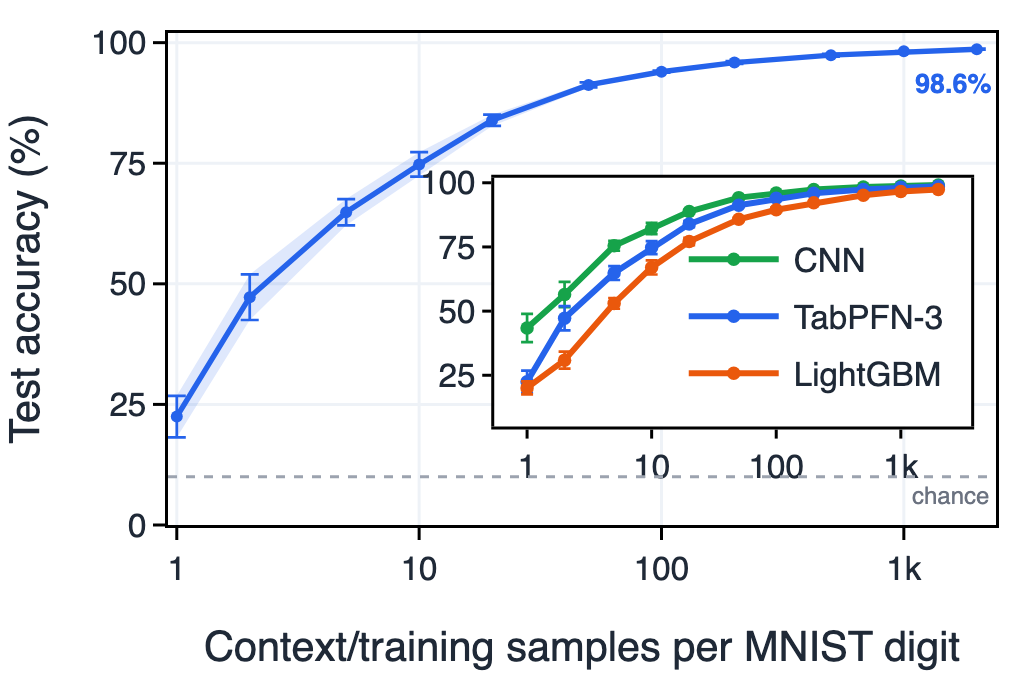}
\caption{
Test accuracy of TabPFN v3 on MNIST from $784$ raw pixels, as a function of
in-context/training examples per digit (blue), evaluated on a fixed
$N_{\text{test}}=5{,}000$-image test set. Error bars indicate $\pm$one standard deviation over resampling the in-context/training examples.
\textbf{Inset}: test accuracy of a simple CNN (green) and LightGBM (orange), a
gradient-boosted tree model, trained on the identical splits, test set, and
number of training sampling repetitions.
    \label{fig:mnist1}
}
\end{figure}

We test how far TabPFN v3 carries a task it was never designed for: classifying the ten MNIST digits. Each $28\times28$ image is handed to TabPFN v3 as a flat row of $784$ raw pixel intensities (Figure \ref{tab:mnist_as_table}), with no normalization, augmentation, or
dimensionality reduction. Classification is purely in-context: the training examples (support set) are loaded as context and a label is produced in a single forward pass, without any gradient updates.

All models draw from a single balanced split: a held-out test set of
$N_{\text{test}}=500$ images per digit ($5{,}000$ total) is sampled first and kept fixed, and the
$N_{\text{train}}$-per-digit learning set is drawn from the disjoint remainder.  The learning set is used as the context for TFM, or as the training set for other methods tried for comparison.  For the sake of conciseness, this distinction will often be kept implicit below, and we will use the word `training' to mean either in-context learning or gradient-based training, depending on the model used. 

We vary $N_{\text{train}}$ between 1 and $2{,}000$ ($10$ to $20{,}000$ in-context rows). Each point in
Fig.~\ref{fig:mnist1} is the mean test accuracy over repeated draws of the support (training) set,  and
the error bars show $\pm$one standard deviation ($\sigma$) over those repetitions.

From the $10\%$ chance level, raw pixels alone take TabPFN v3 from $22.5\%$ at one
example per digit to $98.6\%$ at $2{,}000$, with most of the gain bought in the
first $\sim100$ examples and the resampling variance shrinking from $\pm4\%$ to
$\pm0.1\%$ as the context grows.
For comparison, the inset shows two models that
\emph{train} on the same examples, splits, and repetition budget: a simple CNN and LightGBM, a gradient-boosted tree ensemble with regularization scaled to the training size (architectures and training details in Appendix~\ref{sec:appendix_mnist}).
The ordering $\text{CNN} > \text{TabPFN v3} > \text{LightGBM}$ holds across the
whole curve, with the widest gap between TabPFN v3 and the top-performing CNN in the few training example regime.

We note that TabPFN has no notion of the two-dimensional structure of the image, unlike a CNN, which is designed to exploit the image structure.  In addition, the CNN is trained via gradient descent over multiple epochs (Appendix \ref{sec:appendix_mnist}).  Thus, it is rather remarkable that TabPFN using in-context learning can reach accuracies of similar order as the CNN.

\section{Differentiating French and German words}
\label{sec:nlp}

\begin{figure}[t]
\includegraphics[width=0.97\columnwidth]{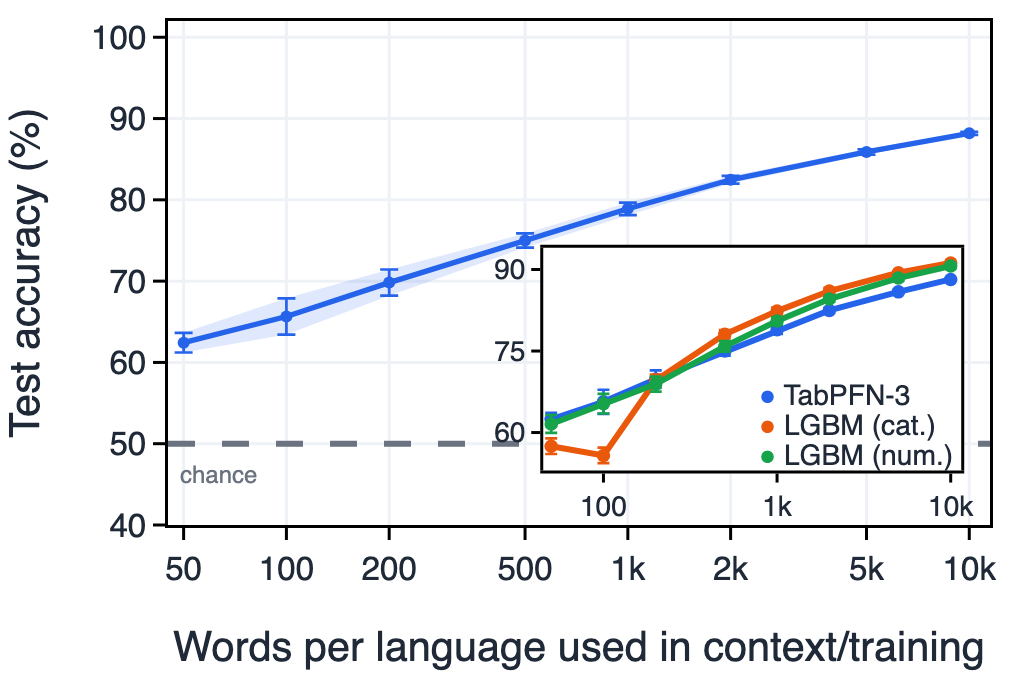}
\caption{
Test accuracy on distinguishing French from German words from their first $15$ letters, as a function of in-context/training words per language, evaluated on a fixed $N_{\text{test}}=4{,}000$-word test set. The main panel shows in-context TabPFN v3 (blue). The shaded band and error bars indicate $\pm$one standard deviation over $10$ in-context/training resamplings. \textbf{Inset}: the same TabPFN v3 curve against tuned LightGBM (green and orange), on identical word sets for train/context and test.
    \label{fig:germanfrench1}
}
\end{figure}

We next apply TabPFN v3 to a task further from its design: deciding whether a word is French or German. Each word is handed to TabPFN v3 as a flat $15$-column ordinal vector of its raw letters ($\text{a}\!\to\!1,\dots,\text{z}\!\to\!26$, zero-padded), with accents and umlauts stripped and no linguistic features. The words are drawn from \texttt{wordfreq}~\cite{robyn_speer_wordfreq}; preprocessing details are in Appendix~\ref{sec:appendix_nlp}. Classification is again purely in-context, in a single forward pass without gradient updates.

All models draw from a single fixed split: a held-out test set of $N_{\text{test}}=4{,}000$ words ($2{,}000$ per language) is sampled first and kept disjoint from every training draw. We vary the training set from $N_{\text{train}}=50$ to $10{,}000$ words per language (log-scaled $x$-axis). Each curve in Fig.~\ref{fig:germanfrench1} is the mean test accuracy over $10$ training resamplings. At a given seed, all models train on identical words, so differences reflect the model, not the sample. Error bars, and the shaded band for TabPFN v3, show $\pm$one standard deviation over the $10$ seeds.

From a $50\%$ chance level, the ordinal letters alone take TabPFN v3 from $62.4\%$ at $N_{\text{train}}=50$ to $88.2\%$ at $N_{\text{train}}=10{,}000$. For comparison, LightGBM trains on the same $15$ columns under two readings: \emph{numeric}, passing the ordinals verbatim, and \emph{categorical}, treating each column as a $27$-level factor so letters can be grouped into unordered subsets rather than ordered thresholds. The two carry identical information and differ only in the available split geometry (tuning in Appendix~\ref{sec:appendix_nlp}). The ordering crosses over: TabPFN v3 leads in the few-shot corner ($N_{\text{train}}\le100$), where the categorical encoding collapses to near a single class, but both tuned LightGBM encodings overtake it from $N_{\text{train}}\ge500$, the categorical reading ending about $3$ points ahead at $N_{\text{train}}=10{,}000$.

Since neither model is particularly designed for this task, this example is instructive for comparing the performance of in-context learning of a foundation model with that resulting from actual training of a model without pretraining.

\section{Tiny ImageNet}
\label{sec:tiny_imagenet}

\begin{figure}[t]
\includegraphics[width=0.97\columnwidth]{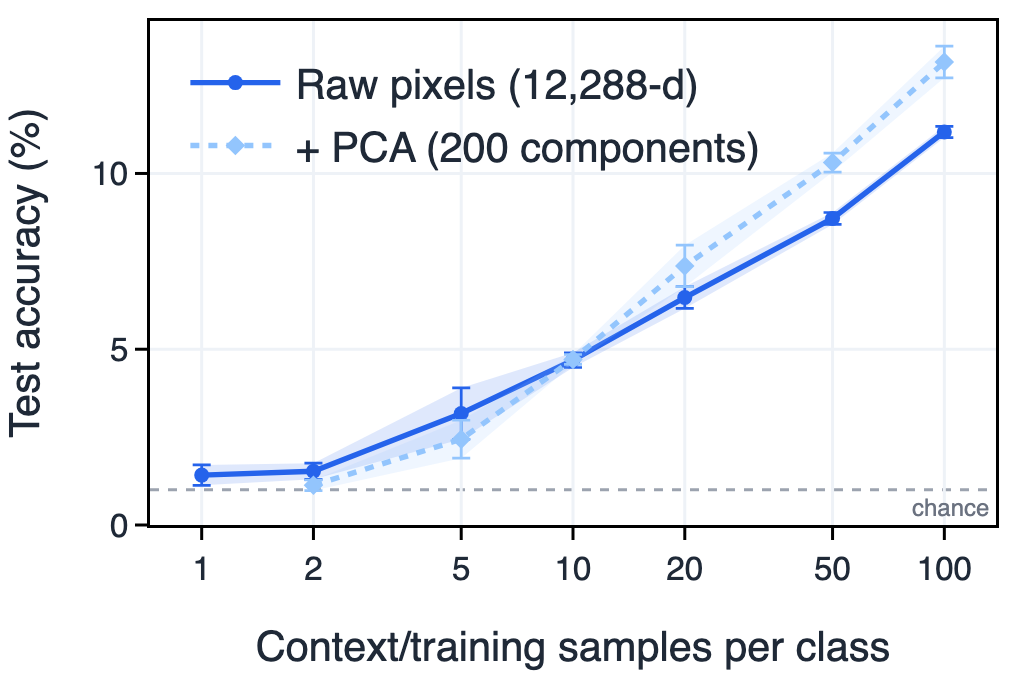}
\caption{
Test accuracy of TabPFN v3 on the $100$-class Tiny ImageNet subset from $12{,}288$ raw pixels (blue), as a function of training examples per class, evaluated on a fixed $N_{\text{test}}=5{,}000$-image test set. The light-blue curve reduces the pixels to $200$ PCA components before TabPFN v3, so that every ensemble estimator covers all features. Error bars indicate $\pm$one standard deviation over resampling the training data.
    \label{fig:tinyimagenet}
}
\end{figure}

Our hardest task pushes TabPFN v3 onto natural images: a $100$-class Tiny ImageNet subset, where each $64\times64$ RGB image is handed over as a flat row of $12{,}288$ raw pixel values, with no normalization or resizing. As before, classification is purely in-context, in a single forward pass without gradient updates. Unlike MNIST, the $12{,}288$ features exceed what TabPFN v3's ensemble samples at its default size, so we also test a variant that first reduces the pixels to 200 features via a principal component analysis (PCA) (details in Appendix~\ref{sec:appendix_tiny_imagenet}).

The test set of $N_{\text{test}}=50$ images per class ($5{,}000$ total) is drawn from the official validation split and the $N_{\text{train}}$-per-class training set from the disjoint training split. We vary $N_{\text{train}}$ between 1 and $100$ ($100$ to $10{,}000$ in-context rows). Each point in Fig.~\ref{fig:tinyimagenet} is the mean test accuracy over repeated training draws, and the error bars show $\pm$one standard deviation over those repetitions.

From the $1\%$ chance level, raw pixels alone take TabPFN v3 from $1.4\%$ at one example per class to $11.2\%$ at $100$. This performance is well above chance, but far short of the near-ceiling accuracies on MNIST. 

Here the raw pixels are not fully covered. Seeing all $12{,}288$ would require $\lceil 12{,}288/200\rceil=62$ ensemble estimators, but TabPFN v3 autoscales the ensemble only up to its default cap of $32$, so about half the pixels go unsampled (Appendix~\ref{sec:appendix_tabpfn_config}). Reducing the pixels to $200$ principal components via PCA instead lets every estimator see all features. In the few-shot corner this brings no gain, as the components estimated on the small training set are too noisy to help. From $N_{\text{train}}\ge20$ the picture changes. The PCA variant draws level with the raw pixels and then pulls ahead, reaching about $2$ points higher accuracy at $100$ examples per class. That covering every feature recovers only about two points reveals the real limit here: not the ensemble size but the absence of a spatial prior, which raw high-dimensional natural images require.

This example highlights the size limitation of current-generation TFMs.  It will be interesting to see how future TFMs, with larger size capabilities, fare with such types of data.

\section{Conclusion}
\label{sec:conclusion}

We applied TabPFN v3 to three tasks outside its tabular domain, namely MNIST and Tiny ImageNet image classification and French-versus-German word recognition. Each example was handed to the model with minimal preprocessing as a flat row of raw features. For each task we swept the in-context training set size across orders of magnitude, from the few-shot regime up to tens of thousands of examples. Throughout, TabPFN's accuracy rose monotonically and did not saturate at the largest context we tried.

The data-rich end is the more surprising. TabPFN is pretrained for small tables, with defaults that assume at most ten thousand samples, yet on the larger sweeps we push it to tens of thousands of in-context rows with no degradation. On MNIST it reaches $98.6\%$ accuracy from raw pixels, just behind a trained CNN and ahead of gradient-boosted trees. On the language task it reaches $88\%$ accuracy from bare letter codes, though tuned trees edge past it. Tiny ImageNet is the exception, as raw pixels never reach an accuracy significantly over $10\%$.

The few-shot end is where TabPFN's zero-tuning nature pays off. From a single example per digit on MNIST it already reaches $22\%$ accuracy, well above the $10\%$ chance level, and on the language task it stays competitive with a per-size-tuned LightGBM, all in a single forward pass with no tuning of its own. Only once data is plentiful does the tuned LightGBM pull clearly ahead.

One might ask why graphical data (MNIST or Tiny ImageNet) can be meaningfully addressed using tabular foundation models.  A partial answer might lie in the pretraining strategy --- the models are exposed not only to real-world datasets but also to (synthetically generated) causal structures.  These examples of causal relationships, generated by sampling random directed acyclic graphs (DAGs), are meant to provide the foundation models with the capability to detect causal structure among the entries in tables.  Looking back at Figure \ref{tab:mnist_as_table}, the essence of handwritten digit recognition can be described as the causal relationship between the 784 pixel values on the one hand, and the label on the other.  The foundation model appears to be able to learn this from a few context samples, without the need for handcrafted techniques like convolution.  The root of this success is presumably the pretraining on causal structures.  This suggests that tabular foundation models might be used for a wide variety of tasks that they have not been trained for, since understanding causality (i.e., understanding world models and relationships between disparate types of data) lies at the heart of prediction and all forms of machine learning.  

The present work implies that tabular foundation models in particular, and causality-awareness in general, hold great promise for a wide range of tasks, well beyond the originally envisaged use cases.

%
%
%

\bibliography{references}

\appendix
\section{Experimental Details}

\subsection{TabPFN v3 configuration}
\label{sec:appendix_tabpfn_config}

All TabPFN results use model v3 library defaults with a single override, \texttt{ignore\_pretraining\_limits=True}. It disables the input-size checks that would otherwise raise an error: v3's soft limits are $500$ features, $10{,}000$ in-context samples. Every experiment trips at least one of these. The feature limit is exceeded by MNIST ($784$ pixel columns) and Tiny ImageNet ($12{,}288$), while the sample limit is exceeded by all three experiments at their largest in-context sizes.

TabPFN handles wide tables by ensembling: each of \texttt{n\_estimators} forward passes sees a subsample of at most \texttt{max\_features\_per\_estimator} columns ($200$ for the v3 default preprocessors), drawn without replacement from a single pool that is shuffled once and consumed round-robin across estimators, so a column cannot repeat until the whole pool has been drawn once. \texttt{auto\_scale\_n\_estimators=True} (default) raises \texttt{n\_estimators} from its default of $8$ toward $\lceil n_{\text{features}}/200\rceil$ so every column is seen by at least one estimator, capped at $32$.

For MNIST ($\lceil 784/200\rceil=4$) and French/German ($\lceil 15/200\rceil=1$) the cap never binds and every column is covered. Tiny ImageNet would need $\lceil 12{,}288/200\rceil=62$ estimators, so the cap leaves only $32\times200=6{,}400$ feature slots for $12{,}288$ columns. Because the round-robin pool guarantees no repeats before exhaustion exactly $6{,}400$ of the $12{,}288$ raw pixels are sampled, leaving $48\%$ of the pixels never seen by any estimator.

\subsection{MNIST}
\label{sec:appendix_mnist}

All models share one balanced split of the MNIST dataset: the $N_{\text{test}}=5{,}000$-image test set ($500$ per digit) is sampled first and fixed, and the $N_{\text{train}}$-per-digit training set drawn from the disjoint remainder. Resampling repetitions follow a size-adaptive schedule, from $10$ at $N_{\text{train}}\le5$ down to $2$ at $N_{\text{train}}\ge1{,}000$.

TabPFN v3 receives the $784$ columns with no preprocessing and classifies purely in-context (training rows loaded as context, each label from a single forward pass), using the v3 configuration of Appendix~\ref{sec:appendix_tabpfn_config}, under which the auto-scaled ensemble covers all $784$ columns.

\begin{figure}[tbhp]
\includegraphics[width=\columnwidth]{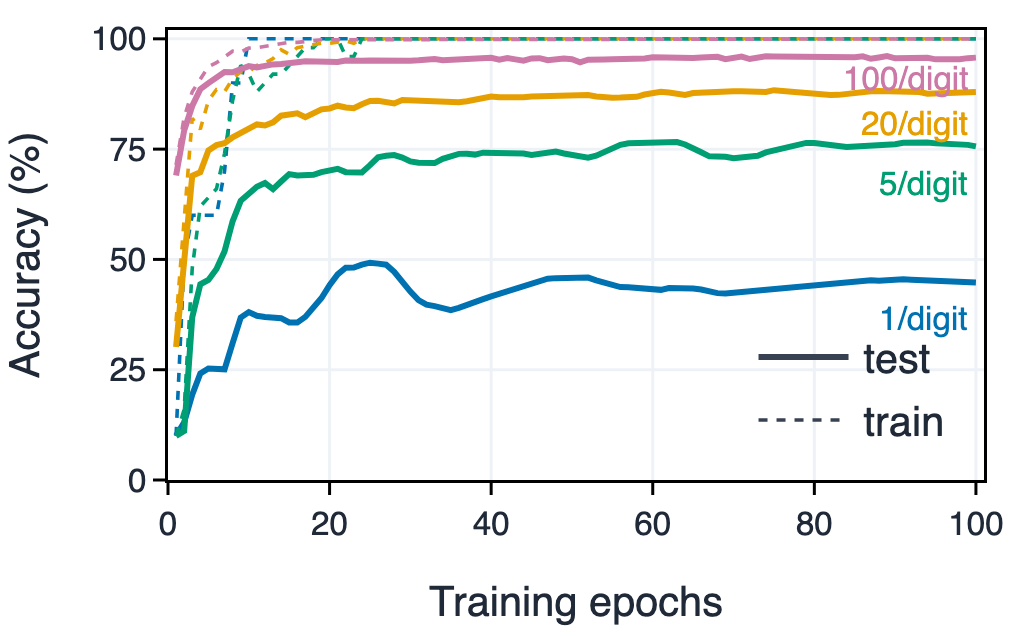}
\caption{
CNN test (solid) and train (dotted) accuracy versus training epochs for single runs on MNIST, for four training-set sizes $1, 5, 20, 100$ examples per digit (from bottom to top). The size of the test set is fixed at $5{,}000$ images. 
    \label{fig:minist_cnn_epochs}
}
\end{figure}

The CNN has two convolutional layers ($32$ and $64$ channels, $3\times3$ kernels, each with ReLU and $2\times2$ max-pooling) and two fully connected layers ($128$ hidden units, then $10$ outputs), $420$k parameters total. Unlike TabPFN v3, its pixel values are scaled to $[0,1]$ and standardized with the MNIST statistics (mean $0.1307$, standard deviation $0.3081$). It trains with Adam (learning rate $10^{-3}$, batch size $128$) and dropout $0.5$ before the output layer, for a fixed $100$ epochs at every training set size. We fix the epoch count rather than early-stop because test accuracy plateaus once the training set is fit and does not degrade with more epochs (Fig.~\ref{fig:minist_cnn_epochs}): the overfitting is benign, reaching $100\%$ train accuracy at every size while test accuracy still improves with $N_{\text{train}}$.

LightGBM sees the same $784$ columns, with regularization scaled to the training size rather than searched: \texttt{num\_leaves} from $4$ to $31$, \texttt{min\_child\_samples} from $1$ to $20$, \texttt{colsample\_bytree} at $0.5$ for $N_{\text{train}}\le2$ else $0.8$, and row bagging (\texttt{subsample}$=0.8$) only once $N_{\text{train}}\ge5$; boosting is fixed at $300$ trees, learning rate at $0.05$. Scaling \texttt{min\_child\_samples} is essential: its default of $20$ forbids any split once a leaf cannot hold $20$ rows, so at $1$ to $2$ examples per digit ($\le20$ rows) the model cannot split and collapses to chance.

\subsection{Differentiating French from German words}\label{sec:appendix_nlp}

Words are drawn from the \texttt{wordfreq} frequency lists (German, French).
Each word is lowercased and normalized to the bare Latin alphabet: German
umlauts and eszett are transliterated as digraphs
($\text{\"a}\!\to\!\text{ae}$, $\text{\"o}\!\to\!\text{oe}$,
$\text{\"u}\!\to\!\text{ue}$, $\text{\ss}\!\to\!\text{ss}$), and all remaining
diacritics are removed via Unicode's normalization form compatibility decomposition ($\text{\'e}\!\to\!\text{e}$, $\text{\c c}\!\to\!\text{c}$, etc.). The remaining letters are then a,b,$\dots$,z. Words are deduplicated within each language. Normalized words common in both languages are removed from both lists to guarantee unambiguous labels. This procedure yields $78{,}169$ German and $71{,}757$ French words. Each word is ordinal-encoded into $15$ integer columns ($\text{a}\!\to\!1,\ldots,\text{z}\!\to\!26$). Unused positions are padded with $0$. Words longer than $15$ letters are truncated. The truncation length $15$ is the $95$th-percentile of the word length in the cleaned data set.

We deliberately strip French accents rather than retain them. Diacritics such as \'e, \`e, and \c c occur only in French and would provide a near-trivial discriminative shortcut; removing them forces both models to classify from Latin-letter orthography alone, making the reported accuracies a lower bound on language separability.

Two consequences of this normalization should be noted. First, truncation is
asymmetric: $7.5\%$ of German words exceed $15$ characters versus only $0.6\%$
of French words, both because German forms longer words through compounding and
long derivational suffixes, and because the umlaut/eszett transliteration itself
lengthens German words by one character each. The presence or absence of padding therefore weakly encodes word length and  correlates with the German label. Second, accent stripping merges otherwise distinct forms (e.g., \`a/a, o\`u/ou, s\^ur/sur) and contributes to the cross-language collisions removed during deduplication (e.g., French \emph{repr\'esentation} $\to$
\emph{representation}, identical to German \emph{Representation}).

TabPFN v3 receives the $15$ ordinal columns with no further preprocessing, using the v3 configuration of Appendix~\ref{sec:appendix_tabpfn_config}. Here $\lceil 15/200\rceil=1$, so a single estimator already covers every column.

LightGBM is tuned independently at each training size and encoding. For each configuration we run \texttt{sklearn}'s
\texttt{RandomizedSearchCV} ($40$ trials) with stratified $5$-fold
cross-validation on the training set only. The held-out test set is never seen during tuning. The search covers tree capacity (\texttt{num\_leaves}
$\in\{7,15,31,63\}$, \texttt{max\_depth} $\in\{-1,4,8\}$), leaf granularity
(\texttt{min\_child\_samples} $\in\{2,5,10,20,40\}$), regularization
(\texttt{reg\_alpha}, \texttt{reg\_lambda}), sampling
(\texttt{colsample\_bytree}, \texttt{subsample}), and the
learning-rate/boosting-round trade-off (\texttt{learning\_rate}
$\in\{0.02,0.05,0.1\}$, \texttt{n\_estimators} $\in\{100,300,600\}$); best
parameters are then refit on the full training set and scored once on the test
set. Crucially the space includes small \texttt{min\_child\_samples}: LightGBM's
default of $20$ forbids any split once a leaf cannot hold $20$ rows, collapsing
to the majority class at small $N_{\text{train}}$. The selected configurations follow the
expected regularization-to-capacity progression with training size: at
$N_{\text{train}}\!\le\!100$ the search prefers shallow, small-leaf, small-step trees
(median \texttt{num\_leaves}$\approx$11--15, \texttt{max\_depth}$\le$6,
\texttt{learning\_rate}$\le$0.05), whereas at $N_{\text{train}}\!\ge\!5000$ it selects
deeper, higher-capacity models at the maximum boosting budget
(\texttt{max\_depth}$=-1$, \texttt{n\_estimators}$=600$,
\texttt{learning\_rate}$=0.1$).

\subsection{Tiny ImageNet}
\label{sec:appendix_tiny_imagenet}

Tiny ImageNet \cite{le2015tiny} contains $200$ classes of $64\times64$ RGB images ($500$ training and $50$ validation images per class). We fix a random $100$-class subset and encode each image as its $64\times64\times3=12{,}288$ raw pixel values, with no normalization or resizing. All models share one balanced split over this subset: the $N_{\text{test}}=5{,}000$-image test set ($50$ per class) is drawn from the official validation split and the $N_{\text{train}}$-per-class training set from the official training split, so the two are disjoint by construction. Resampling repetitions follow a size-adaptive schedule, from $10$ at $N_{\text{train}}\le2$ down to $3$ at $N_{\text{train}}\ge20$.

TabPFN v3 receives the $12{,}288$ columns with no preprocessing and classifies purely in-context, using the v3 configuration of Appendix~\ref{sec:appendix_tabpfn_config}. Here the feature count far outruns the ensemble's reach: the $32$-estimator cap leaves about $48\%$ of the pixels unsampled. We therefore also report a variant that reduces the raw pixels to $200$ PCA components, fit on the training set of each run. As $200$ matches the per-estimator feature budget, every estimator then covers all components and no columns are dropped.

Concretely, each image is treated as a raw-pixel row vector $x\in\mathbb{R}^{12{,}288}$, and the $N_{\text{train}}$ training images are stacked into $X\in\mathbb{R}^{N_{\text{train}}\times12{,}288}$. Writing $\bar{x}$ for the column-wise mean of $X$ and
\[
\Sigma=\frac{1}{N_{\text{train}}-1}(X-\bar{x})^\top(X-\bar{x})\in\mathbb{R}^{12{,}288\times12{,}288}
\]
for its empirical covariance, PCA-200 keeps the eigenvectors $v_1,\dots,v_{200}$ of $\Sigma$ corresponding to the $200$ largest eigenvalues. These are obtained once per run on the training set only. By stacking the eigenvectors as columns of $V=[v_1,\ldots,v_{200}]$, every test image is then represented by its projection
\[
V^\top(x-\bar{x})\in\mathbb{R}^{200}.
\]
This PCA requires at least $200$ rows to extract $200$ nonzero eigenvectors, so the $1$-per-class point is omitted for this variant.

\end{document}